%% file: icra_workshop_extended.tex
\documentclass[10 pt, conference]{ieeeconf}

\IEEEoverridecommandlockouts                

\let\labelindent\relax
\input{header.tex}

\title{\LARGE \bf Novel Algorithms for Smoothly Differentiable \\ and Efficiently Vectorizable Contact Manifold Construction}

\author{\normalsize
        Onur Beker,
        A.\ Ren\'e Geist,
        Anselm Paulus,
        Georg Martius, \\
        University of Tübingen
        }

\begin{document}
\maketitle



\begin{abstract}
Generating intelligent robot behavior in contact-rich settings is a research problem where zeroth-order methods currently prevail. Developing methods that make use of first/second order information about rigid-body dynamics in the presence of contact holds great promise in terms of increasing the solution speed and computational efficiency. The main bottleneck in this research direction is the difficulty in obtaining gradients and Hessians that are actually useful for numerical optimization, due to pathologies in all three steps of a common simulation pipeline: i) collision detection, ii) contact dynamics, iii) time integration. This abstract proposes a method that aims to address the collision detection part of the puzzle, via a novel pipeline designed from scratch with smooth (i.e. twice) differentiability and massive vectorizability on GPUs as the main priorities. This is in contrast to standard collision detection routines that are instead optimized for runtime on CPUs and minimal memory footprint, but do employ logic and control flow that hinder differentiability and vectorization. The proposed pipeline consists of the following contributions: i) highly expressive and compute efficient SDF representations, ii) differentiable broad-phase and narrow-phase routines that use these representations to generate vertex-SDF and edge-SDF contacts, iii) a differentiable routine for convex decomposition based contact blending.
\end{abstract}

\section{Introduction and Related Work} \label{intro}
\blfootnote{*\textit{An earlier version of this manuscript is accepted for presentation at the ICRA 2026 Workshop on Contact-Rich Control and Representation. The present document contains late-breaking additions in preparation for the workshop presentation.}}
Our exposition proceeds as follows. We first briefly describe how existing collision detection pipelines based on convex primitive decomposition work, which are employed as the de-facto standard by all commonly used simulators in robotics \cite{todorov2012mujoco, drake, coumans2016pybullet}. We then outline existing approaches for making such convex primitive based routines smoothly differentiable. Finally, we list the reasons why such approaches still fall short of providing a satisfactory, generally applicable solution to differentiable collision detection.

The standard non-differentiable process employed by commonly used simulators to generate a contact manifold (i.e. a properly distributed set of contact points) between two surfaces can be abstracted into the following steps \cite{erleben2018methodology, gregorius2015robust}:
\begin{itemize}[leftmargin=*, label={\color{ourblue}\textbf{\textbullet}}]
    \item Converting each non-convex surface into a mesh with well-distributed faces using a computational geometry algorithm \cite{jakob2015instant, Corman:2025:RSP, oh2025pamo}, and then decomposing it into a set of convex meshes \cite{wei2022coacd}. All of the remaining steps operate on pairs of these convex meshes.
    \item Running standard collision detection routines like GJK+EPA \cite{gilbert1988fast, montaut2022collision, van2001proximity} or SAT \cite{gregorius2013sat} to obtain information such as a pair of \say{witness points} (the end-points of the largest penetration distance) or closest mesh facets.
    \item Building a contact manifold that sufficiently represents the entire volume/area of intersection between surfaces.
\end{itemize}
The last step of going from two witness points to a manifold is needed to stabilize simulations in pathological settings such as completely parallel faces (e.g. two boxes stacked on top of each other). Alternative ways of implementing it are \cite{gregorius2015robust}:
\begin{itemize}[leftmargin=*, label={\color{ourblue}\textbf{\textbullet}}]
    \item Incrementally building it across simulation time-steps by maintaining a buffer of witness points based on distance heuristics about when to add or remove from the buffer.
    \item Building a contact manifold from scratch every time-step (i.e. \say{one-shot}) by: i)~running GJK+EPA to identify the penetration vector and tangential axes to it, and ii)~inducing small rotations on the two geometries multiple times around the tangential axes and re-running GJK+EPA each time.
    \item Building a contact manifold one-shot by applying polygon clipping algorithms \cite{sutherland1974reentrant} to the closest mesh-facets.
\end{itemize}

Approaches for obtaining smoothed gradients for the witness point positions between two convex primitives can be grouped in two: analytical smoothing and randomized smoothing. A prominent example of the former is the work of \citet{tracy2023differentiable}, which solves for the growth distance \cite{ong1996growth} between pairs of a representative set of convex primitives (including convex meshes) by formulating them as cone-constrained convex programs. Gradients through this convex program are then obtained via implicit differentiation and smoothed via modulating the log-barrier coefficient of the underlying interior-point solver \cite{tracy2024differentiability}. A prominent example of the randomized smoothing approach is the work by \citet{montaut2023differentiable}, which operates by: i)~sampling $M$ deviations around a nominal kinematic configuration, 
ii)~running GJK and EPA algorithms \cite{gilbert1988fast} $M$ times to get separation distances, 
iii)~getting a zeroth-order estimate of the gradient via the score function estimator \cite{williams1992simple}. While both methods can robustly generate useful gradients for the witness point positions between two convex primitives, we argue that the underlying convex primitive based paradigm itself is pathological for differentiable contact simulation, as we elaborate on next.

The first pathology of the convex primitive based formulation is that witness points are conceptually not well defined (i.e. non-unique) for configurations that involve parallel contacts between planar faces, which are very common in contact simulations since they often constitute stable points for the dynamics (e.g. a box on a plane). For the analytical smoothing approach, this means in such pathological configurations (where the non-smoothed witness point optimization problem is not strictly convex), the uniqueness comes from the (strictly convex) log-barrier term, which needs to be sufficiently flat for the gradients to remain physically plausible, thus leading to slow convergence. For the randomized smoothing approach, this means the estimator has large variance at such configurations, since witness points can jump drastically in arbitrary directions based on the sampled deviation. 
The second pathology is that a single witness points pair is not sufficient for stable simulation, and only addressing the differentiability of the convex-convex witness point routine is therefore insufficient. For the subsequent step of building a contact manifold from witness points, maintaining a buffer couples gradients across timesteps which is unsuitable for optimal control, and making polygon clipping routines differentiable in a computationally efficient manner is non-trivial (due to branching code and non-smooth operations such as top-K selection) \cite{paulus2025hard} and remains an open problem. The most commonly used option is therefore inducing incremental rotations along tangential axes and re-running the convex-convex routine to build a manifold. The main problem is that there is no standard, mathematically principled way of picking the tangential axes uniquely (since they are only defined up to a rotation along the normal direction), in a way that is guaranteed to change smoothly across timesteps.
A third and final pathology is that in the absence of a broad-phase routine for filtering primitive pairs without running the complete convex-convex routine, the convex primitive based formulation is inefficient. This is because for two surfaces decomposed into $N$ and $M$ primitives each, $N \times M$ convex-convex checks are required. Therefore only addressing the narrow-phase convex-convex routine is insufficient since the broad-phase routine can still create jumps in the gradient (due to primitive pairs activating and deactivating). 

All of these considerations motivate us to design a collision detection routine from scratch with differentiability and vectorizability as the primary concerns (which constitutes the main contribution of this paper), inspired by the analogous routines of barrier-based contact simulation approaches \cite{Li2020IPC}. We opt for this approach rather than trying to make existing routines differentiable, because these standard routines were conceived within a computer graphics, video game design, or computational science and engineering context with speed (i.e. of a single simulation instance on CPU) and minimal memory footprint as the main priorities. 

\section{Background: Smooth Approximations to Non-Smooth Operators} \label{sec:smooth_operators}
The proposed framework extensively uses smooth approximations of non-smooth operators such as comparisons and clipping. This section therefore aims to provide a highly compressed summary for completeness. A much more detailed exposition can be found in the SoftJAX library \cite{paulus2026softjax}.

The sigmoid function $\sigma(x) = \left(1 + \exp(-x)\right)^{-1}$ can be used to smoothly approximate a comparison operator $\llbracket x > a \rrbracket: \mathbb{R} \to [0, 1]$ (notated with an Iverson-Bracket) via $\sigma((x-a)/\softness)$. 
The $\text{relu}(x)$ function, which projects a number onto the positive line-segment, can be smoothly approximated using a softplus function $\text{s}_+(x) = \softness \log(1 + \exp(x / \softness))$. The clipping operation $f_\text{clip}(x, x_\text{min}, x_\text{max}) = x_\text{min} + \text{relu}(x - x_\text{min}) - \text{relu}(x - x_\text{max})$ can in turn be softened into a softclip operation $s_\text{clip}(x, x_\text{min}, x_\text{max})$ by replacing $\text{relu}$ operations with $\text{s}_+$. The scalar sigmoid and softplus functions can be generalized to operate on vectors $\bm x \in \mathbb{R}^D$ to obtain the softmax and realsoftmax (i.e. logsumexp) functions respectively, notated as $s_\text{argmax}(\bm x)_i = \exp(x_i / \softness)/\sum_{j=1}^D \exp(x_j / \softness)$ and $\text{LSE}(\bm x) = \softness\log[\sum_{i=1}^D\exp(x_i / \softness)]$. Note that $\bm s_\text{argmax}(\bm x) \in \mathbb R^D$ is a vector of probability masses that approximates an argmax operation, whereas $\text{LSE}(\bm x)\in \mathbb R$ is a scalar that corresponds to a soft selection of the maximum component of $\bm x$. 

\section{Methods: Surface Representations}
In the proposed framework, collisions happen between one surface represented with a signed distance field (SDF), and the vertices and edges of another surface that is represented with a mesh. Symmetry is achieved by simply running the routine again with these representation type assignments (i.e. SDF or mesh) transposed. Therefore, the smooth differentiability of the entire routine relies on constructing SDFs that are smoothly differentiable functions, which is the focus of this section. The proposed framework provides two alternative ways of constructing such SDFs to suit different needs: i) as a superposition of analytical primitives called \say{extruded plane-superquadric intersections} (XPSQ) , ii) as 3D splines (which themselves are superpositions of cubic polynomial basis functions). The benefits of the former primitive-based approach are that: i) it is very memory efficient, and ii) it easily admits randomized alterations in a way that preserves surface semantics (e.g. for procedural generation or domain randomization); but has the downside that it needs to be constructed by hand for each new surface since no automated decomposition method yet exists that can read a mesh and return a list of XPSQs that approximate it. The latter spline-based approach is less memory efficient and harder to randomize, but can automatically be constructed from arbitrary meshes and is therefore easier to work with.

\subsection{Primitive-Based SDF Representation}
This section elaborates on the proposed \say{extruded plane-superquadric intersection} (XPSQ) primitive, extending the line of work by \citep{beker2025, beker2026smoothly}, in particular on the superquadric (SQ) primitive.
At a high-level, the XPSQ primitive analytically solves for the SDF associated with the volume traced by sweeping a superquadric (potentially intersected with $N$ half-spaces) along a 1D quadratic spline curve embedded in 3D. This enables describing geometries such as the handle of a cup with a single primitive, which would not be possible via convex decomposition \cite{wei2022coacd} or an SQ decomposition \cite{liu2023marching}.


\begin{figure}[h]
\begin{center}
\includegraphics[width=\columnwidth]{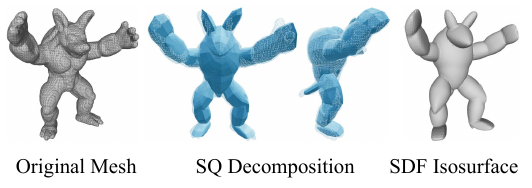}
\end{center}
\vspace{-3mm}
\caption{An armadillo geometry approximated as the smooth union of 18 superquadric (SQ) SDFs, illustrating the highly expressive nature of SQ primitives that from the foundation of XPSQ primitives.}
\label{fig:sq1}
\vspace{-2.5mm}
\end{figure}
\noindent\textbf{1) Superquadrics.}
Superquadrics (SQ) are a family of surfaces that subsume basic shape primitives such as cubes, cylinders, and ellipsoids. 
An SQ in its canonical form is defined by the unit level set the level set $f(\bm x)=1$ of an \emph{inside--outside} implicit function $f:\mathbb{R}^3\rightarrow\mathbb{R}$ \cite{liu2023marching}:
\begin{align}
f(\mathbf{x}) =
\left(
    \left(\frac{x}{a_x}\right)^{\frac{2}{\epsilon_2}}
    +
    \left(\frac{y}{a_y}\right)^{\frac{2}{\epsilon_2}}
\right)^{\frac{\epsilon_2}{\epsilon_1}}
+
\left(\frac{z}{a_z}\right)^{\frac{2}{\epsilon_1}}
\end{align}
where $\epsilon_1$ controls the shape's ``roundness'', $\epsilon_{2}$ adjusts the shape's ``pointedness'', and $a_x$ scales the $x$-dimension. In total there are 11 parameters: 6 for 3D pose, 3 for per axis scales, 2 for roundness and pointedness.

\noindent\textbf{2) Differentiable Combinations of SDFs.} 
Any two SDFs $\phi_1(\bm x),\phi_2(\bm x): \mathbb{R}^3 \to \mathbb{R}$ (e.g. representing a half space, SQ, or XPSQ) can be combined through \emph{union} $\phi_3=\min(\phi_1,\phi_2)$, \emph{intersection} $\phi_3=\max(\phi_1, \phi_2)$, or \emph{subtraction} $\phi_3=\max(\phi_1, -\phi_2)$ operations. Smoothly differentiable versions of these operations are obtained through the use of the logsumexp operator \cite{sdf_algebra, smoothmin}:
\begin{align}
    \phi_\cup(\bm x) &= -\text{LSE}([-\phi_1(\bm x), -\phi_2(\bm x)]) \hspace{2mm} \text{(smooth union)}, \label{eq:soft_union} \\
    \phi_\cap(\bm x) &= \text{LSE}([\phi_1(\bm x), \phi_2(\bm x)]) \hspace{2mm} \text{(smooth intersection)}, \label{eq:smooth-intersection}\\
    \phi_{\ominus}(\bm x) &= \text{LSE}(\phi_{+}(\bm x), -\phi_{-}(\bm x)) \hspace{2mm} \text{(smooth subtraction)}.
\end{align}

\noindent\textbf{3) PSQ: Combining Half Spaces with SQs.} 
The SDF of a half-space is defined as 
$\phi_N = \bm x \cdot \bm n + h$,
where $\bm n$ denotes the plane normal and $h$ the offset from the origin. Through the smooth intersection \eqref{eq:smooth-intersection} of an SQ and a single half space, most of the geometric primitives provided by common simulation frameworks can be defined, including boxes, ellipsoids, cylinders, elliptic cones, square pyramids, and tetrahedrons (although our framework and software implementation does not assume any restrictions on the number of half spaces intersecting an SQ). We refer to the primitive obtained by intersecting $N$ half spaces $\mathbf{P} \in \mathbb{R}^{N \times 4}$ ($3$ parameters for $\bm n$ and $1$ for $h$) and an SQ 
as a \say{plane-SQ intersection} (PSQ). 



\begin{figure}[h]
    \centering
    \includegraphics[width=0.8\linewidth]{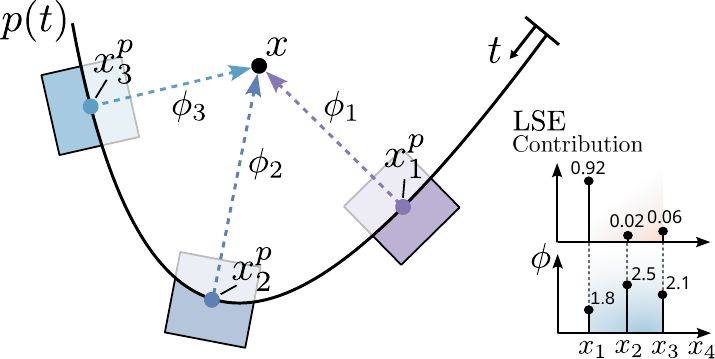}
    \caption{The SDF of the XPSQ primitive is evaluated by: (i) analytically projecting a point $\bm x$ onto a spline $p(t)$, (ii) transforming the PSQ SDF expression such that the surface is moved to coincide with each of the three projection points, and (iii) obtaining the SDF for $\bm x$ via the smooth-minimum of the three transformed PSQ SDFs.}
    \label{fig:spline_sdf}
\end{figure}

\noindent\textbf{4) XPSQ: Tracing a Spline with a PSQ.} 
The XPSQ primitive represents the volume obtained by tracing a PSQ along a 1D quadratic spline curve embedded in 3D.
Such a quadratic spline curve admits a parameterization using $t \in [0, 1]$ and three control points $\bm{p}_{1:3}$ as follows:
\begin{equation}
    \bm{p}(t) = (1-t)^2\bm{p}_1 + 2t(1-t)\bm{p}_2 + t^2 \bm{p}_3.
\end{equation}
The data-structure that represents an XPSQ primitive therefore maintains the following information: i) a quadratic spline $\bm p(t)$, ii) a smooth function $\mathbf{R}(t) \in \text{SO(3)}$ that prescribes the rotational pose of the PSQ as it traces the spline (e.g. the Frenet frame associated with the spline \cite{spivak_frenet_serret}), iii) smooth functions $\epsilon_{\{1, 2\}}(t)$ and $a_{\{x, y, z\}}(t)$ that prescribe the shape and scale parameters of the SQ part of the PSQ, iii) $\mathbf{P}(t) \in \mathbb{R}^{N \times 4}$ that prescribes the plane normal (expressed in the PSQ base frame) and shift for the $N$ half spaces within the PSQ. 

\begin{figure}[h]
    \centering
    \includegraphics[width=\linewidth]{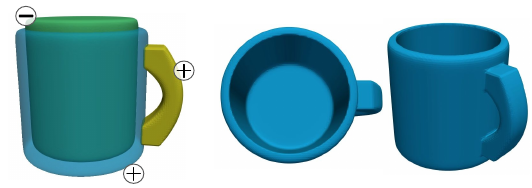}
    \caption{Combining XPSQ primitives via SDF union, intersection, and subtraction operations provides an efficient and highly expressive way to represent SDFs of challenging surfaces, such as a cup.}
    \label{fig:xpsq_combination}
\end{figure}
\noindent\textbf{5) The SDF Associated with an XPSQ.}
To evaluate the SDF of the XPSQ for a given point $\bm{x}$, one first projects it onto the spline to obtain the projection point $\bm{p}(t^*)$, and then evaluates the SDF of the PSQ at that point (as prescribed by $\mathbf{R}(t^*), \epsilon_{\{1, 2\}}(t^*), a_{\{x, y, z\}}(t^*), \mathbf{P}(t^*)$).
As was also noted by \cite{Li25RAL}, this projection can be obtained by minimizing $\| \bm x - \bm{p}(t)\|^2$, which is a fourth order polynomial in $t$. Taking its gradient and setting it to zero in turn gives a cubic equation in $t$, which can be solved analytically via Cardano's formula \cite{artin2011algebra}.
The solution is not unique in general, as for some constellations there are multiple points on the spline with equally minimal distance to $\bm{x}$, e.g. when the spline is wrapped around $\bm{x}$.
In Cardano's formula, these cases are distinguished by the sign of the discriminant $\Delta\in\mathbb{R}$, which is a function of the control points, and the computation of the roots branches depending on it.
This branching leads to a discontinuity, which we soften by always treating both branches simultaneously and merging the results.
In the negative branch we use the negative projected discriminant $\Delta^- = -s_+(-\Delta)\in\mathbb{R}^-$ to obtain the unique real root $t^-$ of the cubic.
In the positive branch, we use the positive projected discriminant $\Delta^+ = s_+(\Delta)\in\mathbb{R}^+$, yielding three real roots
$(t_1^+, t_2^+, t_3^+)$.
Note that in both cases we apply a soft clipping operation to keep the roots in the range $(0,1)$, as Cardano's formula can return roots outside of this range.
Next, we always return three roots as the convex combination $\bm t^* \in \mathbb{R}^3$ of the two root sets:
\begin{equation}
    \bm t^* = \llbracket \Delta < 0 \rrbracket * [t^-, t^-, t^-] + \llbracket \Delta > 0 \rrbracket * [t_1^+, t_2^+, t_3^+] 
\end{equation}
where the brackets are smoothed as described in S.\ref{sec:smooth_operators}.
Then, we get the PSQs prescribed at all three roots using $\mathbf{R}(t^*_{i})$, $\bm x^*_{i}$, $ \epsilon_{\{1, 2\}}(t^*_{i})$, $a_{\{x, y, z\}}(t^*_{i})$, and $\mathbf P(t^*_{i})$ with $i \in \{1, 2, 3\}$. 
The associated SDFs for the three PSQs are $\{\phi(\bm x^*_1), \phi(\bm x^*_2), \phi(\bm x^*_3)\}$, and we finally return their smooth-minimum $\phi_{\cup}(\bm x) = -\text{LSE}([-\phi_1(\bm x), -\phi_2(\bm x), -\phi_3(\bm x)])$ as shown in Fig.\ref{fig:spline_sdf}.

\subsection{Spline-Based SDF Representation}
\begin{figure}[h]
\begin{center}
\includegraphics[width=\columnwidth]{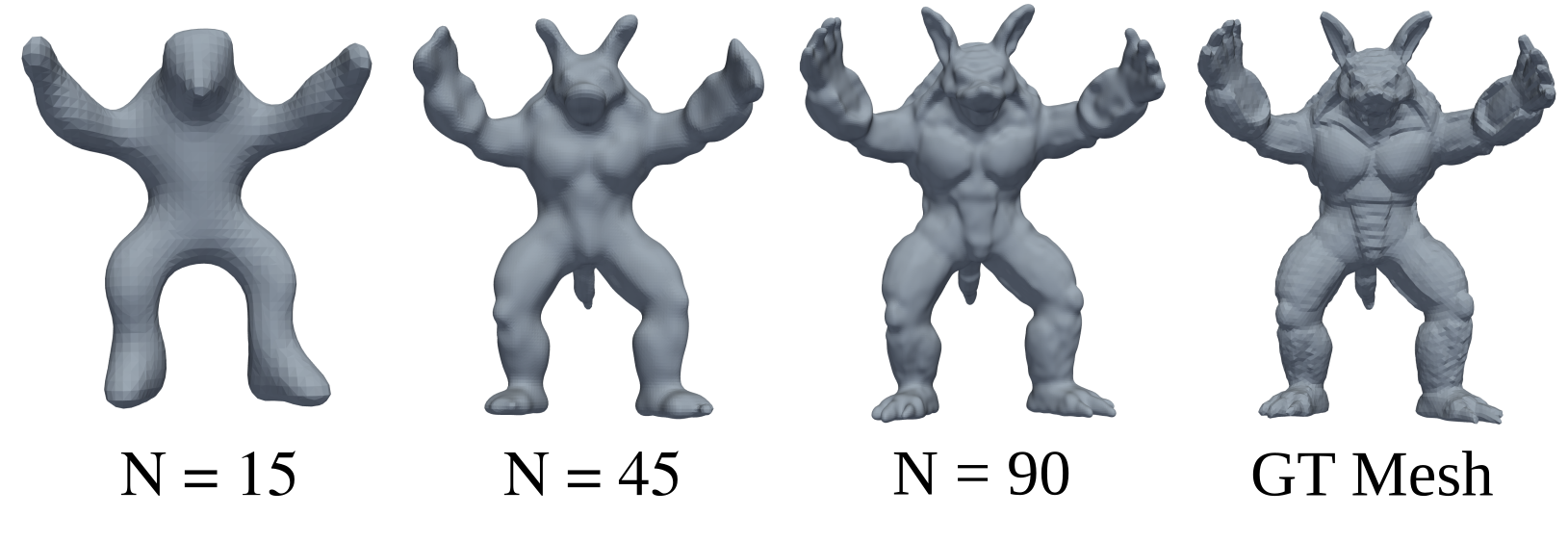}
\end{center}
\vspace{-3.5mm}
\caption{Spline primitives provide an efficient and automated way of obtaining smooth approximations to the SDF of any input mesh. The voxel density $N$ per axis determines the amount of smoothing, with larger numbers providing closer (but sharper) approximations.}
\label{fig:spline1}
\vspace{-5mm}
\end{figure}
This section provides a short background on how 3D cubic splines that map $\mathbb R^3 \to \mathbb R$ are constructed\footnote{Please refer to \cite{habermann2007multidimensional, floater2023Spline, lai2007spline} for a more in-depth treatment.}, and elaborates on how they can be used to represent SDFs and surface normals.

\noindent\textbf{1) Cubic Splines in 3D.} 
In 1D, standard cubic splines are built as a sum of twice-differentiable polynomial basis functions $u: \mathbb{R} \to \mathbb{R}$ of the following form \cite{habermann2007multidimensional}:
\begin{align}
    u(t) = 
    \begin{cases}
    4 - 6|t|^2 + 3|t|^3 \quad &\text{for} \ |t| \leq 1  \\
    (2 - |t|)^3  \quad &\text{for} \ \ 1 \leq |t| \leq 2 \  \\ 
    0 \quad &\text{for} \ 2 \leq |t|
    \end{cases}
\end{align}

Given the values $\{f_i\}_{i=0}^N$ of a function $f: [a, b] \to \mathbb{R}$ evaluated at the corners of $N$ intervals splitting $[a, b]$ into equal pieces of length $h$ each, the cubic spline interpolation in 1D is given by the sum $s(t) = \sum_{k=1}^{N + 3} c_k u_k(t)$, using a superposition of $N+3$ shifted basis functions defined as $u_i(t) = u(\frac{t - a}{h} - k + 2)$. The coefficients $c_k$ are in turn obtained by solving the linear system defined by the equations $s(t_i) = f_i$.

Given the values $\{f_{i, j, k}\}_{i, j, k=0}^{N_x, N_y, N_z}$ of a function $f: \mathbb{R}^3 \to \mathbb{R}$ evaluated at the corners of a 3D voxel grid with $(N_x, N_y, N_z)$ voxels along each axis, a 3D spline interpolation is obtained as $s(x, y, z) = \sum_{k_x=0}^{N_x+3}\sum_{k_y=0}^{N_y+3}\sum_{k_z=0}^{N_z+3} \ c_{k_x, k_y, k_z} \ u_{k_x}(x) u_{k_y}(y) u_{k_z}(z)$, using products of the form $u_{k_x}(x) u_{k_y}(y) u_{k_z}(z)$ to define 3D basis functions from 1D basis functions. The coefficients $c_{k_x, k_y, k_z}$ are again obtained by solving the linear system defined by the equations $s(x_i, y_j, z_k) = f_{i, j, k}$.

An important property of spline interpolation is that, because the basis function $u(t)$ has compact support ($u(t) = 0$ for $t \notin [-2, 2]$), once the coefficients are obtained, evaluating the spline at arbitrary points is very fast and memory efficient. In 3D, only $64$ of the terms in the sum $s(x,y,z)$ are nonzero for any point $(x, y, z)$, and in general there are only $4^D$ nonzero terms for a spline in $D$ dimensions. Therefore, one only needs to identify the indices of the $64$ nonzero terms via integer rounding operations, after which they can be evaluated and summed with little computational cost.

\noindent\textbf{2) Using 3D Splines to Represent SDFs and Normals.} The ground truth signed distance and surface normal functions defined by a mesh can be evaluated at any 3D point $\bm x$ by: 
\begin{itemize}[leftmargin=*, label={\color{ourblue}\textbf{\textbullet}}]
    \item Finding the closest point $\text{proj}(\bm x)$ to $\bm x$ on the mesh surface.
    \item Performing an occupancy check to determine whether $\bm x$ is inside or outside the mesh as $\text{sign}(\bm x) \in \{-1, 1\}$.
    \item Computing the SDF via $\phi(\bm x) = \text{sign}(\bm x)\|\bm x - \text{proj}(\bm x)\|$ and the surface normal via $\bm n(\bm x) = \text{sign}(\bm x)\frac{\bm x - \text{proj}(\bm x)}{\|\bm x - \text{proj}(\bm x)\| + \epsilon}$.
\end{itemize}

Evaluating the ground truth SDF and surface normal functions (that are only once-differentiable) along the corners of a voxel grid and performing a 3D cubic spline interpolation gives a twice-differentiable approximation (where the amount of smoothing is controlled by the voxel grid density). The reason surface normals are not obtained from the gradient of the SDF interpolation is because they also need to be twice-differentiable for ensuring the smooth differentiability of the entire simulation pipeline.

\section{Methods: Contact Manifold Construction}
Given the previously described primitive-based and spline-based approaches that provide an efficient way to represent SDFs of complex surfaces, the next step is using them to create contact manifolds between colliding surfaces in a differentiable and vectorizable manner. In the proposed routine, vertices and edges of a mesh collide with an SDF. This happens in three consecutive stages: i) broad-phase, ii) narrow-phase, and iii) blending-phase. In the broad-phase, all vertices and edges that are guaranteed to lie further away than a given distance threshold from the opposing surface are identified and filtered via checks that are more efficient than running the subsequent narrow-phase computations on them. The narrow-phase is where the actual vertex-SDF (V-SDF) and edge-SDF (E-SDF) collisions happen for those edges and vertices that were not filtered. Finally, in the blending-phase, information such as penetration depths, contact normals and Jacobians associated with all V-SDF and E-SDF contacts are fused together using precomputed group IDs based on a convex decomposition, to significantly reduce the total number of contact points that will be passed to the subsequent contact dynamics stage of the overall simulation pipeline. The following sections elaborate on these proposed broad-phase, narrow-phase, and blending-phase routines. Although broad-phase comes before narrow-phase in the overall pipeline, it is covered after narrow-phase for better exposition.
\subsection{Preparing SDF and Mesh Representations}
Before delving into the details of the proposed collision detection pipeline, let us first describe step-by-step how the collision meshes and collision SDFs are obtained starting from an arbitrary input mesh:
\begin{itemize}[leftmargin=*, label={\color{ourblue}\textbf{\textbullet}}]
    \item The input mesh is processed to obtain the collision SDF. For the primitive-based approach, this is done either through modelling by hand or using an automated decomposition approach such as the method of \cite{liu2023marching}. For the spline-based approach, this is done by computing ground truth SDF and surface normal values with respect to the input mesh and solving for the spline coefficients.
    \item Once the collision SDF is obtained, it is converted into an iso-surface mesh using marching cubes. The conversion from the input mesh to the collision SDF and then back into an iso-surface mesh happens because the collision SDF does not have to exactly match the input mesh (since it is smoothed to be twice-differentiable).
    \item The iso-surface mesh is processed with the re-meshing method of \cite{jakob2015instant} to obtain a reduced number of faces that have high isotropy and in a way that the edges and corners align well with the sharp features of the underlying geometry. This results in the collision mesh.
\end{itemize}
Since the collision mesh is obtained from simplifying the iso-surface of the collision SDF, these two representations can differ slightly. If it is desirable to obtain an exact match between them, one first runs the proposed collision detection pipeline to obtain contact points on the collision mesh, and then projects these contact points $\bm p$ onto the corresponding collision SDF isosurface via $\text{proj}(\bm p) = p - \phi(\bm p) \bm n(p)$.

\subsection{Narrow-phase}
Narrow-phase is where contact points, penetration depths, and contact normals are computed for those V-SDF and E-SDF collisions that are not filtered during broad-phase. V-SDF collisions amount to treating every vertex as a contact point and evaluating the SDF and surface normal functions at these points to assign a penetration depth and a contact normal. E-SDF collisions involve selecting a suitable intersection point among all points on an edge, which is handled via sphere-tracing. Given the straight-forward nature of V-SDF collisions, the rest of this section focuses on E-SDF contacts.

\noindent\textbf{1) Finding Edge-SDF Intersection Points.} Points $\bm e$ on an edge between corners $\vertex_I, \vertex_{II} \in \mathbb{R}^3$ can be parameterized as $\edge(\alpha) = \vertex_I + \alpha \bm e_t $ with $\bm e_t = \frac{\vertex_{II} - \vertex_I}{\|\vertex_{II} - \vertex_I\|}$ and $0 \leq \alpha \leq \|\vertex_{II} - \vertex_I\|$. This means the maximal penetration point between the edge $e$ and any SDF $\phi$ can hypothetically be found via $\min_{\alpha} \ \phi(\bm e(\alpha))$ subject to $0 \leq \alpha \leq \|\vertex_{II} - \vertex_I\|$. One can then obtain smooth gradients through this constrained optimization problem by employing log barrier smoothing and applying the implicit function theorem to the resulting KKT conditions \cite{tracy2023differentiable, tracy2024differentiability}. That being said, there are a number of pathologies associated with employing such an edge-SDF collision routine:
\begin{itemize}[leftmargin=*, label={\color{ourblue}\textbf{\textbullet}}]
    \item The optimization problem is non-linear, meaning the solution and its gradients obtained via implicit differentiation would still exhibit jumps when the problem switches from one local optimum to another as parameters are varied.
    \item Unrolling the solver steps (e.g. Newton's method in 1D) and applying automatic differentiation is also not an option since it is computationally expensive. This is because: i) computing Hessians is expensive, and ii) the number of Newton iterations needs to be large (empirically $\geq10$).
    \item As shown in \cite{beker2026smoothly}, such procedures are inefficient for vectorization, because a vectorized iterative solver applied to a batch of problems will only finish as quickly as the slowest-converging problem in the batch.
\end{itemize}

\begin{figure}
    \centering
    \includegraphics[width=0.7\linewidth]{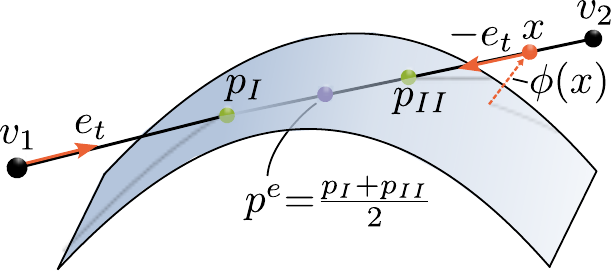}
    \caption{Intersection points $p_I$ and $p_{II}$ between the SDF $\phi(x)$ of one surface and the edge $v_1-v_2$ of the other surface are determined via sphere tracing. At every sphere tracing step, the current intersection point candidate $x$ is moved towards the SDF by $\pm \llbracket \phi(x) > 0 \rrbracket \phi(x) e_t$. The $\llbracket \phi(x) > 0 \rrbracket$ term ensures that if $v_1$ or $v_2$ already lie inside the surface, they are not moved by sphere tracing.}
    \label{fig:e_s}
    \vspace{-5mm}
\end{figure}
 
\noindent\textbf{2) Collision Detection with Sphere Tracing.} 
Therefore, we devise a custom edge-SDF collision routine based on sphere-tracing \cite{hart1996sphere}. Given an initial point $\bm p_0$ and a ray direction $\bm n$, sphere-tracing is an algorithm to find ray-SDF intersections (widely considered to be the most efficient way of doing so even for very complex surfaces \cite{crane2005ray}). It proceeds via updates of the form $\bm p_{k+1} = \bm p_{k} + \llbracket \phi(x) > 0 \rrbracket \phi(x) \bm n$. The proposed edge-SDF routine then proceeds as follows:
\begin{itemize}[leftmargin=*, label={\color{ourblue}\textbf{\textbullet}}]
    \item Sphere trace the edge corners $\vertex_I$ and $\vertex_{II}$ along $\bm e_t$ and $-\bm e_t$ respectively to obtain $\bm p_I$ and $\bm p_{II}$. We have empirically found three iterations to be more than sufficient for convergence.
    \item Clip $\bm p_I$ and $\bm p_{II}$ within the edge boundaries via a soft-clip operation, and return the midpoint $\bm p^e = (\bm p_I + \bm p_{II}) / 2$.
\end{itemize}
This process is illustrated in \Cref{fig:e_s}.
Naturally, it is guaranteed to succeed only in cases when an edge penetrates the SDF only once or never. We argue that this is sufficient, because if there exists an edge that penetrates an SDF more than once, this indicates that the mesh tessellation is not at a density adequate for the complexity of the surfaces involved in the collision, and should hence be increased.

After the E-SDF intersection point $\bm p^e$ is found using the proposed sphere-tracing routine, it is treated as a contact point and the SDF and surface normal functions are then evaluated to assign penetration distances and contact normals.

\subsection{Broad-phase}
\noindent\textbf{1) Narrow-phase Modifications to Accommodate Differentiable Broad-phase.} The goal of the broad-phase is to identify those V-SDF and E-SDF collisions that are guaranteed to produce contact points with a penetration distance larger than a distance threshold $d_\text{max} \geq 0$, so that they can be omitted from the subsequent narrow-phase for efficiency. The reason we do not set $d_{max}=0$ as default is because it is preferable to devise contact models that can exert forces and return gradients at a distance \cite{mordatch2012discovery, pang2023global, paulus2025hard}, an application that the proposed collision detection pipeline also aims to facilitate. To be able to omit collisions further than $d_\text{max} \geq 0$ in a manner that preserves twice-differentiability, we define the following mollification operations to modify contact information returned by the narrow-phase:
\begin{itemize}[leftmargin=*, label={\color{ourblue}\textbf{\textbullet}}]
    \item Employing a transition function $f_t: \mathbb{R} \to [0, 1]$ (e.g. a piece-wise polynomial) that is non-increasing, twice-differentiable, and satisfies $f_t(d) = 0$ for $d > d_\text{max}$ and $f_t(d) = 1$ for $d \leq d_\text{max} - \epsilon$ for a given small bandwidth $\epsilon \geq 0$.
    \item Redefining the penetration depth $d_c$ of a contact point $\bm p$ as $d_c(\bm p) = f_t(\phi(\bm p))\phi(\bm p) + [1 - f_t(\phi(\bm p))]d_\text{max}$ such that $d_c$ saturates to the value $d_\text{max}$ with increasing SDF $\phi(\bm p)$.
    \item Redefining the contact normal as $\bm n_c(\bm p) = f_t(\phi(\bm p)) \ \bm n(\bm p)$, and the contact Jacobian as $\mathbf J_c(\bm p) = f_t(\phi(\bm p)) \ \mathbf J (\bm p)$, such that contact points that are further away than $d_\text{max}$ from the opposing surface do not exert any contact force (assuming the subsequent contact dynamics solver processes $d_c, \bm n_c, \mathbf J_c$ in a way that properly handles these modifications).
\end{itemize}

\begin{figure}[h]
\begin{center}
\includegraphics[width=\columnwidth]{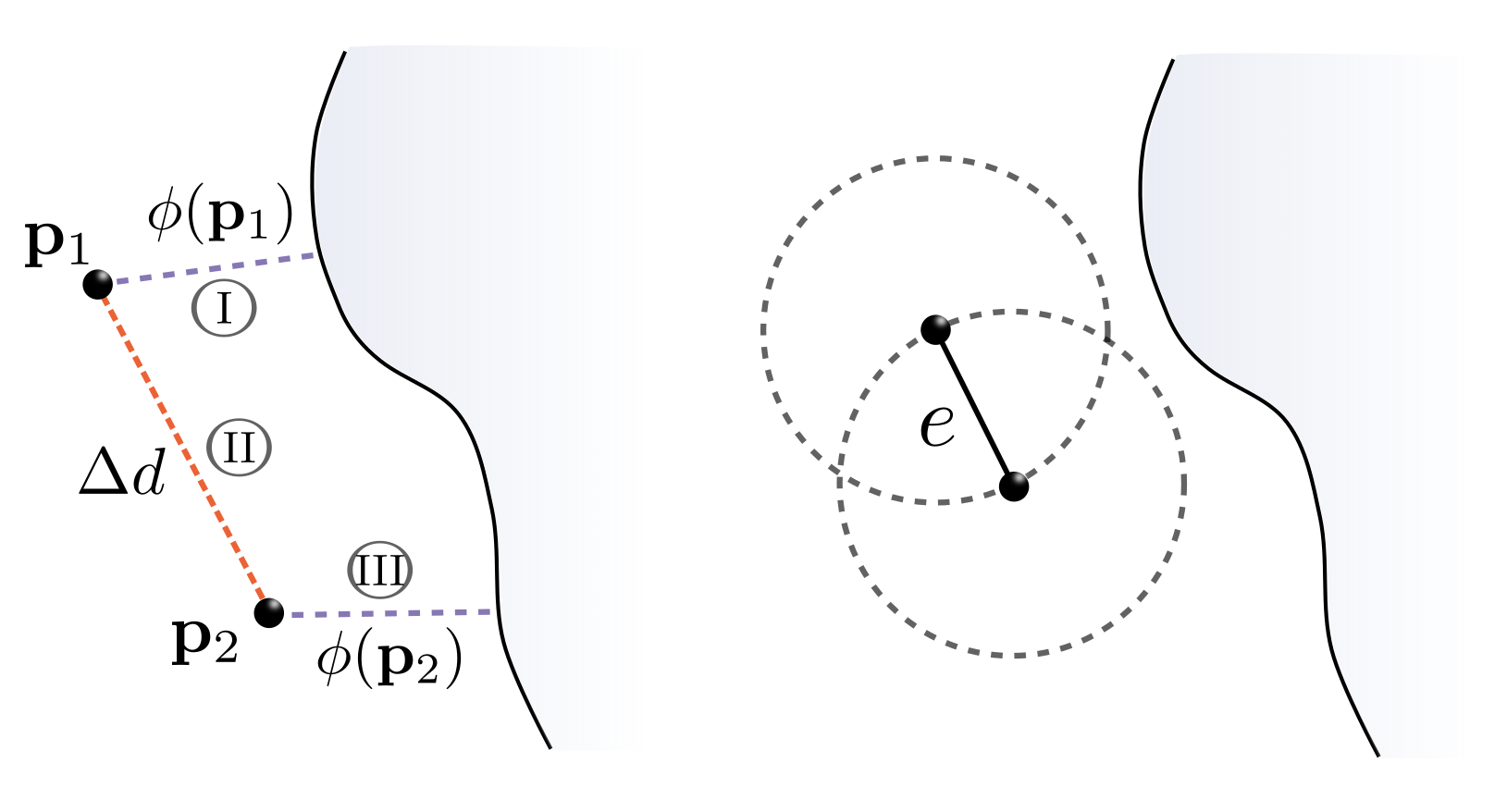}
\end{center}
\vspace{-3mm}
\caption{\textbf{Left}: Among all curves that start at a point $\bm p_1$ and end on a surface, the shortest one is the line segment with length $\|\phi(\bm p_1)\|$ and direction $\nabla \phi(\bm p_1)$. For two points $(\bm p_1, \bm p_2)$ with distance $\Delta d$ in-between, this implies that $\|\phi(\bm p_1)\| \leq \|\phi(\bm p_2)\| + \Delta d$ holds (since the left side is the length of curve I which is the shortest, and right side is the length of curves II-III concatenated). This property can be used to pre-compute SDF values on a voxel grid and use them to filter vertices in broad-phase. \textbf{Right}: An edge with endpoints $(\bm p_1, \bm p_2)$ and length $l$ cannot collide with a surface as long as $\min(\phi(\bm p_1), \phi(\bm p_2) \ ) \geq l$, which can be combined with the previous property (used for vertex filtering) to filter edges in broad-phase.}
\label{fig:broadphase1}
\vspace{-5mm}
\end{figure}

\noindent\textbf{2) Differentiable Broad-phase.} The main enabler of the broad-phase routine is a voxel grid (constructed in the body frame of the surface represented with an SDF) where every voxel stores the value of the SDF function evaluated at its center point. Reading values from this voxel grid is a very fast $\mathcal{O}(1)$ operation. Furthermore, the voxel grid does not need to be duplicated when the contact manifold generation routine is vectorized (since GPUs allow parallel reads from the same memory address), which means storing the voxel-grid does not constitute a memory bottleneck for massive vectorization. 

Let $\bm p$ be a 3D point that is $\Delta d$ away from the center $\bm p_c$ of the voxel within which it falls, with the associated SDF values $\phi(\bm p)$ and $\phi(\bm p_c)$ (where $\phi(\bm p_c)$ is already computed and stored with the voxel grid). As shown in Fig.\ref{fig:broadphase1}, the inequality $\phi(\bm p) + \Delta d \geq \phi(\bm p_c)$ always holds. Given the radius $r$ of the circumsphere of each voxel, this means $\phi(\bm p_c) - r$ can be used as a lower-bound for $\phi(\bm p)$. Therefore V-SDF collisions can be filtered by finding the voxel a vertex falls into and checking whether $\phi(\bm p_c) - r \geq d_\text{max}$ holds. If it does, the V-SDF collision is filtered away and does not participate in the narrow-phase, and instead automatically gets assigned $(d_\text{max}, \bm 0, \mathbf 0)$ as its penetration depth, contact normal and Jacobian.

Let $\bm e$ be an edge with length $l$ and endpoints $(\bm p_1, \bm p_2)$ with corresponding SDF values $(\phi(\bm p_1), \phi(\bm p_2))$. Again using the same arguments from Fig.\ref{fig:broadphase1}, for any point $\bm p$ on the edge, $\phi(\bm p) + l \geq \phi(\bm p_1)$ and $\phi(\bm p) + l \geq \phi(\bm p_2)$ both hold. As in the V-SDF case, this means $\min(\phi(\bm p_1), \phi(\bm p_2)) - r - l$ can be used as a lower-bound for $\phi(\bm p)$, which can again be compared to $d_\text{max}$ to filter away E-SDF collisions and automatically assign $(d_\text{max}, \bm 0, \mathbf 0)$ as the resulting narrow-phase output.

\begin{figure}[h]
\begin{center}
\includegraphics[width=\columnwidth]{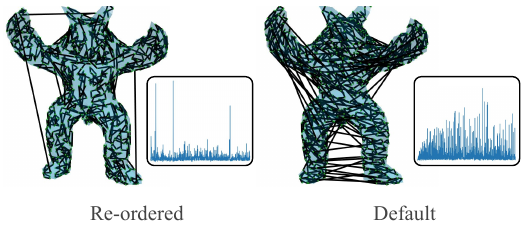}
\end{center}
\vspace{-3mm}
\caption{Vertex re-ordering using a Hilbert curve for serialization, to accelerate broad-phase filtering. The lines connect subsequent vertices. The plots in turn show distances between subsequent vertices in the y axis, and indices themselves in x axis. It can be seen that re-ordering significantly reduces jumps between the 3D positions of vertices that have subsequent indices.}
\label{fig:broadphase2}
\vspace{-1.0mm}
\end{figure}
\noindent\textbf{3) Vertex and Edge Reordering to Optimize Memory Access.} A GPU organizes threads into groups of 32 called warps, and all 32 threads in a warp progress through kernel instructions in lock step fashion as specified by the Single-Instruction Multiple-Threads (SIMT) paradigm \cite{nvidia_cuda_guide}. This means if a kernel contains a control flow branch (e.g. an if/else statement) that is followed by $N$ threads in the warp, the remaining $32 - N$ threads sit idle (a phenomenon named \say{warp divergence}). Minimizing warp divergence means minimizing idle time, making vectorized code run faster.

For the proposed broad-phase and narrow-phase combination, warp divergence is minimized if one maximizes the number of 32 consecutive (i.e. in terms of index order in the array that stores them) vertices/edges that are all accepted or rejected together by broad-phase filtering. Achieving this in turn requires ordering vertices/edges in an array such that if two vertices/edges are close together in terms of 3D Euclidean distance, then their indices are also close together. This is achieved by re-ordering vertices/edges on a mesh based on the traversal order of their projections on a 3D space-filling curve, a standard process (often called \say{serialization}) employed in file sequencing to speed up queries \cite{morton1966computer}. The proposed framework supports serialization using Z-order and Hilbert curves, building on  implementations provided by \cite{wang2017ocnn, wu2024ptv3}.

\subsection{Blending-Phase}
\begin{figure}[h]
\begin{center}
\includegraphics[width=0.85\columnwidth]{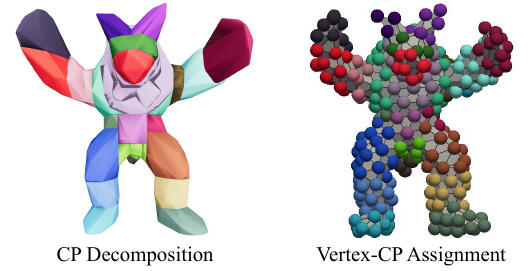}
\end{center}
\vspace{-3mm}
\caption{\textbf{Left}: visualizations of convex primitive (CP) decomposition outputs. \textbf{Right}: vertex CP group ID assignments. Vertices (on the right) that are assigned the same CP group ID are plotted with the same color as the corresponding CP itself (on the left).}
\label{fig:blending1}
\vspace{-1.5mm}
\end{figure}
\noindent\textbf{1) Convex Primitive Assignment.} In preparation for contact blending, meshes of both surfaces are processed via a convex decomposition algorithm \cite{wei2022coacd}, to obtain $N_1$ and $N_2$ convex primitives (CP) respectively. Every CP has an associated list of vertices and edges assigned to it. Vertices are assigned to the CP that they lie inside of, while edges are assigned to every CP they intersect (since meshes are constructed via a computational geometry algorithm \cite{jakob2015instant} to be as isotropic as possible, majority of edges are contained within a single CP and only a very small minority intersects two neighboring CPs). The overall goal of the blending-phase is to reduce the $V_1 + E_1 + V_2 + E_2$ number of contact points to $N_1 + N_2$.

\noindent\textbf{2) Contact Blending (Mechanism).} The proposed (optional) contact blending routine works as follows:
\begin{itemize}[leftmargin=*, label={\color{ourblue}\textbf{\textbullet}}]
    \item Let the index $i \in \{1, \cdots, K\}$ run over all vertex and edge contact points that belong to the same CP, with coordinates $\{\bm p_{c, i}\}_{i=1}^K$ and penetration depths $\{d_{c, i}\}_{i=1}^K$.
    \item A softmax operator is applied on $\{d_{c, i}\}_{i=1}^K$ to obtain blending weights $\{w_{i}\}_{i=1}^K$.
    \item These weights are used to blend all contact information such as penetration depths, surface normals (which are normalized after blending), and contact Jacobians.
\end{itemize}

\noindent\textbf{3) Contact Blending (Justification).} Collisions between two convex polytopes can in most configurations be resolved by identifying the single unique pair of witness points with maximal penetration in between. This approach only becomes problematic within the neighborhood of those pathalogical configurations where a unique pair of witness points with maximal penetration cannot be identified, such as the case of two boxes stacked on top of another with their contacting faces lying perfectly parallel (where there are infinitely many witness point pairs that have the same maximal penetration depth in-between). In such cases, the standard approach is employing a polygon clipping algorithm to explicitly construct the polygon enclosing the area of intersection between two mesh faces touching each other, and use the corners of this polygon as the contact manifold (possibly further selecting a subset of four points that span maximal area) \cite{gregorius2013sat, gregorius2015robust}.\footnote{Note that the proposed sphere tracing based E-SDF contact routine serves an analogous purpose to polygon clipping.}

Such pathological cases that require a contact manifold are characterized by the property that the points on the manifold have nearly identical penetration depths and contact normals. Therefore, it is physically plausible to assume that the contact forces acting on these points are also nearly identical (both in terms of magnitude and direction). In fact, under the standard dissipative spring-damper based models of compliant contact \cite{kurtz2025cenic, kurtz2026inverse, hydroelastic, johnson1987contact, hunt1975coefficient} (that have extensive experimental validation and are thus widely used in commercial multibody simulation tools \cite{mscadams}), these contact forces (that only depend on the contact point depth, normal and velocity) would indeed be identical in static configurations. Given that the contact forces $\bm f_i$ enter the Newton Euler equation as $\mathbf{M}\bm a + \bm b = \tau + \sum_i \mathbf{J}_i^T \bm f_i$ (where the index $i$ runs over points in the contact manifold), and that for nearly identical contact forces $\sum_i \mathbf{J}_i^T \bm f_i \approx (\sum_i \mathbf{J}_i^T)\bm f$ holds, one could argue that the benefits of constructing an entire contact manifold rather than a single witness point pair mainly manifest themselves through the way the contact Jacobian is modified (i.e. determining where exactly the common contact force $\bm f$ should act on the body). And since the contact Jacobian of a free body expressed in body frame is $\mathbf{J}(\bm p_i) = \left[[\bm p_i]_\times \ \ \mathbf I \right] \in \mathbb{R}^{3 \times 6}$ (where $\bm p_i \in \mathbb{R}^3$ is the contact point coordinates in body frame and $[\bm a]_\times \bm b = \bm A \times \bm b$), blending contact points as $\sum_i w_i \bm p_i$ correctly aggregates contact Jacobians as $\mathbf{J}(\sum_i w_i \bm p_i) = \sum_i w_i \mathbf{J}(\bm p_i)$. Additionally, since $\sum_i w_i=1$, the resulting point $\sum_i w_i \bm p_i$ is guaranteed to stay within the convex hull of the contact points $\bm p_i$ being blended.

Note that the proposed blending approach essentially constitutes an inversion of the standard convex primitive based paradigm. In the standard paradigm, convex primitives are first identified and then this decomposition is used to obtain a small number of contact points (since one gets a single contact point per CP pair using the GJK/EPA routine); whereas in the proposed approach contact points are first identified and the convex decomposition is only then utilized to reduce the number of contact points via blending. 

\section{Experiments} \label{experiments}
\begin{figure}[h]
\begin{center}
\includegraphics[width=\columnwidth]{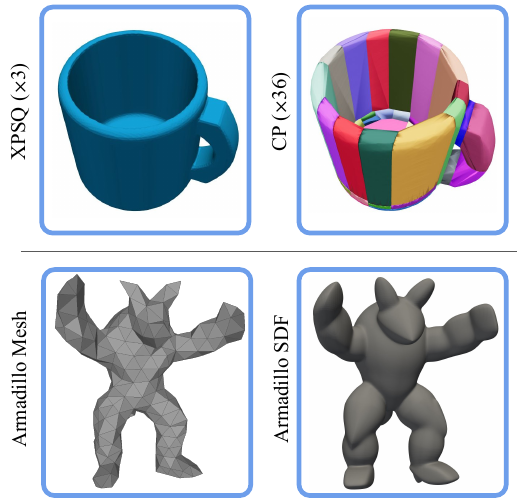}
\end{center}
\vspace{-3mm}
\caption{Examples that demonstrate the expressivity of XPSQs.}
\label{fig:exp}
\vspace{-2.5mm}
\end{figure}
\noindent\textbf{1) Expressivity of the XPSQ Primitive.} The top part of  Fig.\ref{fig:exp} illustrates how a cup can be represented with only 3 XPSQ primitives (i.e. one cylinder subtracted from another, combined with a handle obtained by sweeping a box along a spline). In contrast, convex decomposition using CoACD \cite{wei2022coacd} results in 36 primitives, which, in the absence of a differentiable broad-phase routine, is highly inefficient. The bottom part shows how an armadillo can be efficiently represented with 18 SQs, obtained using the method of \cite{liu2023marching}.

\begin{figure}[th]
\begin{center}
\includegraphics[width=\columnwidth]{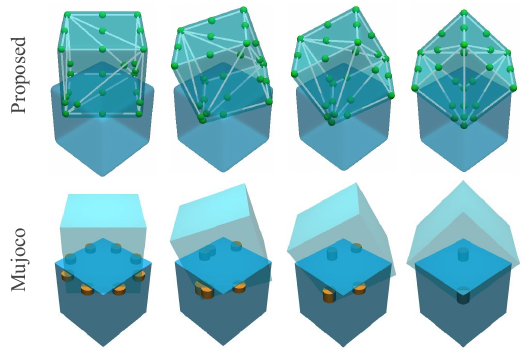}
\end{center}
\vspace{-3mm}
\caption{An example illustrating the behavior of the contact manifold.}
\label{fig:man}
\end{figure}
\noindent\textbf{2) Behavior of the Contact Manifold.} Fig.\ref{fig:man} illustrates the qualitative behavior of the contact manifold compared to the polygon clipping routine employed in Mujoco \cite{todorov2012mujoco}, on an example with two boxes stacked on top of each other. It can be seen that polygon clipping results in non-smooth behavior of the contact points, whereas the proposed edge-SDF contact points adapt smoothly to the surface, adequately sampling the volume of intersection between the colliding surfaces. From this perspective, the proposed routine can be interpreted as a middle ground between polygon clipping and the hydroelastic contact model \cite{hydroelastic}, without needing to explicitly re-mesh the intersection volume like the latter (which is a non-differentiable and non-vectorizable routine).

\begin{figure}[th]
\begin{center}
\includegraphics[width=\columnwidth]{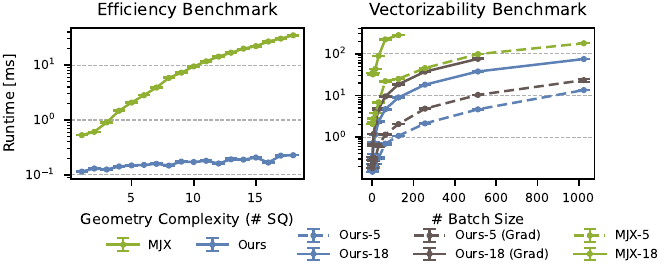}
\end{center}
\vspace{-3mm}
\caption{A benchmark characterizing the computational efficiency of the narrow-phase routine in the absence of broad-phase filtering.}
\label{fig:bench}
\vspace{-3mm}
\end{figure}
\noindent\textbf{3) Computational Efficiency (Narrow-phase).} Fig.\ref{fig:bench} contains plots showing the speed and vectorizability of the proposed narrow-phase routine (implemented in JAX) without broad-phase, compared to the analogous routine of the well-established MJX simulator. The setup involves collisions between two non-convex armadillo geometries across 100 random configurations per-trial. Surfaces are represented with up to 18 SQs each, which is a controlled variable characterizing the effects of geometry complexity on the runtime. The second controlled variable is the vectorization batch size. 
Notice the logarithmic scale and the order of magnitude improvement compared to MJX. 

\begin{figure}[th]
\begin{center}
\includegraphics[width=\columnwidth]{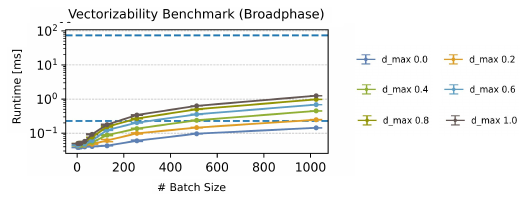}
\end{center}
\vspace{-3mm}
\caption{A benchmark characterizing the computational efficiency of the combined broad-phase and narrow-phase routines.}
\label{fig:bench2}
\vspace{-1mm}
\end{figure}
\noindent\textbf{4) Computational Efficiency (Broad-phase).} Fig.\ref{fig:bench2} characterizes the vectorizability of the combined broad-phase and narrow-phase routines for increasing $d_\text{max}$ values used for filtering, using the exact same armadillo geometry as the previous section. Since broad-phase filtering requires an if/else branch to be placed at the beginning of the narrow-phase routine for early termination, this experiment uses a separate implementation of both phases in NVIDIA-Warp \cite{Macklin_Warp_A_High-performance_2022} (which allows kernels containing branching code to be run without triggering a JIT recompilation unlike JAX). In this experiment, every batch with batchsize $B$ is constructed by translating one armadillo away from the other along the x axis by $B$ different linearly spaced distances between $[1, 2]$. For reference, the axis aligned bounding-box of the armadillo geometry has an extent of $1.38$ units along the x axis, meaning every batch starts from half intersecting bounding boxes (of the two armadillos in collision), and ends with non-intersecting bounding boxes. The horizontal dashed lines in the plot show the runtime of the narrow-phase routine without broad-phase (from the previous section) for batch sizes 1 and 1024, for ease of reference. It can be seen that broad-phase filtering brings another order of magnitude improvement in runtime.

\section{Limitations and Conclusion}
We have presented a collision detection routine that is constructed from scratch with smooth differentiability and vectorizability in mind. This is enabled by three contributions: i) expressive and efficient SDF representations, ii) broad-phase and narrow-phase routines to implement V-SDF and E-SDF collisions, iii) a blending-phase that merges contact points together using convex decomposition based groupings. The main limitation is robotics and control experiments that characterize the usefulness of the gradients obtained from the proposed method, which constitutes the most immediate direction for future work.

\bibliographystyle{IEEEtranN} 
\bibliography{scm_iros}

\end{document}

%% file: header.tex
\usepackage{times}
\usepackage[shortlabels]{enumitem}

\usepackage[utf8]{inputenc} 
\usepackage[T1]{fontenc}    
\usepackage{url}            
\usepackage{booktabs}       
\usepackage{nicefrac}       
\usepackage{microtype}      
\usepackage{adjustbox}      
\usepackage{graphics,color}

\usepackage{dirtytalk}
\makeatletter
\let\NAT@parse\undefined
\makeatother
\usepackage[font=small,labelfont=bf]{caption}
\usepackage{bbm}
\usepackage{bm}
\usepackage{algorithm}
\usepackage[numbers]{natbib}

\usepackage{amsfonts}       
\usepackage{amsmath}       
\usepackage{amssymb}
\usepackage{xspace}
\usepackage{multicol}
\usepackage{epsfig} 
\usepackage{mathptmx} 
\usepackage[makeroom]{cancel}
\usepackage{soul}
\usepackage{epigraph}
\usepackage{stmaryrd}
\usepackage{graphicx}
\usepackage{caption}
\usepackage{tabularx}
\usepackage{tikz}
\usepackage{comment}

\makeatletter
\DeclareRobustCommand\onedot{\futurelet\@let@token\@onedot}
\def\@onedot{\ifx\@let@token.\else.\null\fi\xspace}
\makeatother

\makeatletter
\newcommand*{\addFileDependency}[1]{
  \typeout{(#1)}
  \@addtofilelist{#1}
  \IfFileExists{#1}{}{\typeout{No file #1.}}
}
\makeatother

\newcommand\blfootnote[1]{%
  \begingroup
  \renewcommand\thefootnote{}\footnote{#1}%
  \addtocounter{footnote}{-1}%
  \endgroup
}

\newcommand{\vertex}{\bm{v}}

\newcommand{\edge}{\bm{e}}

\newcommand{\softness}{\tau}

\usepackage{xcolor}
\definecolor{ourblue}{rgb}{0.368,0.507,0.71}    
\definecolor{ourorange}{rgb}{0.881,0.611,0.142} 
\definecolor{ourgreen}{rgb}{0.56,0.692,0.195}   
\definecolor{ourred}{rgb}{0.923,0.386,0.209}    
\definecolor{ourviolet}{rgb}{0.528,0.471,0.701} 
\definecolor{ourbrown}{rgb}{0.772,0.432,0.102}  
\definecolor{ourazure}{rgb}{0.364,0.619,0.782}  
\definecolor{ourolive}{rgb}{0.572,0.586,0.}     
\definecolor{ourgray}{RGB}{102,88,84}           

\definecolor{ourblue2}{RGB}{9,134,223} 
\definecolor{ourdarkblue2}{RGB}{5,97,164} 
\definecolor{ourlightblue2}{RGB}{132,201,250} 

\definecolor{ourorange2}{RGB}{224,90,18} 
\definecolor{ourdarkorange2}{RGB}{160,63,9} 
\definecolor{ourlightorange2}{RGB}{246,175,137} 

\definecolor{ouryellow2}{RGB}{227,213,25} 
\definecolor{ourdarkyellow2}{RGB}{177,166,17} 
\definecolor{ourlightyellow2}{RGB}{242,235,140} 

\definecolor{ourpink2}{RGB}{247,24,139} 
\definecolor{ourdarkpink2}{RGB}{164,4,86} 
\definecolor{ourlightpink2}{RGB}{250,163,207} 

\definecolor{ourgreen2}{RGB}{159,198,52} 
\definecolor{ourdarkgreen2}{RGB}{109,138,30} 
\definecolor{ourlightgreen2}{RGB}{209,228,154} 

\definecolor{ourgray2}{RGB}{124,124,115} 
\definecolor{ourdarkgray2}{RGB}{87,87,81} 
\definecolor{ourlightgray2}{RGB}{194,194,189} 

\usepackage{xr-hyper}
\usepackage[pagebackref,breaklinks,colorlinks,allcolors=ourblue]{hyperref}
\usepackage{cleveref}  %

%% file: icra_workshop_extended.bbl
\begin{thebibliography}{48}
\providecommand{\natexlab}[1]{#1}
\providecommand{\url}[1]{#1}
\csname url@samestyle\endcsname
\providecommand{\newblock}{\relax}
\providecommand{\bibinfo}[2]{#2}
\providecommand{\BIBentrySTDinterwordspacing}{\spaceskip=0pt\relax}
\providecommand{\BIBentryALTinterwordstretchfactor}{4}
\providecommand{\BIBentryALTinterwordspacing}{\spaceskip=\fontdimen2\font plus
\BIBentryALTinterwordstretchfactor\fontdimen3\font minus \fontdimen4\font\relax}
\providecommand{\BIBforeignlanguage}[2]{{%
\expandafter\ifx\csname l@#1\endcsname\relax
\typeout{** WARNING: IEEEtranN.bst: No hyphenation pattern has been}%
\typeout{** loaded for the language `#1'. Using the pattern for}%
\typeout{** the default language instead.}%
\else
\language=\csname l@#1\endcsname
\fi
#2}}
\providecommand{\BIBdecl}{\relax}
\BIBdecl

\bibitem[Todorov et~al.(2012)Todorov, Erez, and Tassa]{todorov2012mujoco}
E.~Todorov, T.~Erez, and Y.~Tassa, ``Mujoco: A physics engine for model-based control,'' in \emph{2012 IEEE/RSJ International Conference on Intelligent Robots and Systems}.\hskip 1em plus 0.5em minus 0.4em\relax IEEE, 2012, pp. 5026--5033.

\bibitem[Tedrake and the Drake~Team(2019)]{drake}
\BIBentryALTinterwordspacing
R.~Tedrake and the Drake~Team, \href{https://drake.mit.edu}{``Drake: Model-based design and verification for robotics},'' 2019.
\BIBentrySTDinterwordspacing

\bibitem[Coumans and Bai(2016)]{coumans2016pybullet}
E.~Coumans and Y.~Bai, ``Pybullet, a python module for physics simulation for games, robotics and machine learning,'' 2016.

\bibitem[Erleben(2018)]{erleben2018methodology}
K.~Erleben, ``Methodology for assessing mesh-based contact point methods,'' \emph{ACM Transactions on Graphics (TOG)}, vol.~37, no.~3, pp. 1--30, 2018.

\bibitem[Gregorius(2015)]{gregorius2015robust}
D.~Gregorius, ``Robust contact creation for physics simulations,'' in \emph{GDC}, 2015.

\bibitem[Jakob et~al.(2015)Jakob, Tarini, Panozzo, and Sorkine-Hornung]{jakob2015instant}
W.~Jakob, M.~Tarini, D.~Panozzo, and O.~Sorkine-Hornung, ``Instant field-aligned meshes,'' in \emph{ACM Transactions on Graphics (Proceedings of SIGGRAPH ASIA)}, vol.~34, no.~6, Nov. 2015.

\bibitem[Corman and Crane(2025)]{Corman:2025:RSP}
E.~Corman and K.~Crane, ``Rectangular surface parameterization,'' \emph{ACM Trans. Graph.}, vol.~44, 2025.

\bibitem[Oh et~al.(2025)Oh, Yuan, Wei, Shi, Xiang, Liu, and Su]{oh2025pamo}
S.~Oh, X.~Yuan, X.~Wei, R.~Shi, F.~Xiang, M.~Liu, and H.~Su, ``Pamo: Parallel mesh optimization for intersection-free low-poly modeling on the gpu,'' in \emph{Computer Graphics Forum}.\hskip 1em plus 0.5em minus 0.4em\relax Wiley Online Library, 2025, p. e70267.

\bibitem[Wei et~al.(2022)Wei, Liu, Ling, and Su]{wei2022coacd}
X.~Wei, M.~Liu, Z.~Ling, and H.~Su, ``Approximate convex decomposition for 3d meshes with collision-aware concavity and tree search,'' \emph{ACM Transactions on Graphics (TOG)}, vol.~41, no.~4, pp. 1--18, 2022.

\bibitem[Gilbert et~al.(1988)Gilbert, Johnson, and Keerthi]{gilbert1988fast}
E.~G. Gilbert, D.~W. Johnson, and S.~S. Keerthi, ``A fast procedure for computing the distance between complex objects in three-dimensional space,'' \emph{IEEE Journal on Robotics and Automation}, vol.~4, pp. 193--203, 1988.

\bibitem[Montaut et~al.(2022)Montaut, Lidec, Petrik, Sivic, and Carpentier]{montaut2022collision}
L.~Montaut, Q.~L. Lidec, V.~Petrik, J.~Sivic, and J.~Carpentier, ``Collision detection accelerated: An optimization perspective,'' \emph{arXiv preprint arXiv:2205.09663}, 2022.

\bibitem[Van Den~Bergen(2001)]{van2001proximity}
G.~Van Den~Bergen, ``Proximity queries and penetration depth computation on 3d game objects,'' in \emph{GDC}, vol. 170, 2001, p. 209.

\bibitem[Gregorius(2013)]{gregorius2013sat}
D.~Gregorius, ``The separating axis test between convex polyhedra,'' in \emph{GDC}, 2013.

\bibitem[Sutherland and Hodgman(1974)]{sutherland1974reentrant}
I.~E. Sutherland and G.~W. Hodgman, ``Reentrant polygon clipping,'' \emph{Communications of the ACM}, vol.~17, no.~1, pp. 32--42, 1974.

\bibitem[Tracy et~al.(2023)Tracy, Howell, and Manchester]{tracy2023differentiable}
K.~Tracy, T.~A. Howell, and Z.~Manchester, ``Differentiable collision detection for a set of convex primitives,'' in \emph{2023 IEEE International Conference on Robotics and Automation (ICRA)}.\hskip 1em plus 0.5em minus 0.4em\relax IEEE, 2023, pp. 3663--3670.

\bibitem[Ong and Gilbert(1996)]{ong1996growth}
C.~J. Ong and E.~G. Gilbert, ``Growth distances: New measures for object separation and penetration,'' \emph{IEEE Transactions on Robotics and Automation}, vol.~12, no.~6, pp. 888--903, 1996.

\bibitem[Tracy and Manchester(2024)]{tracy2024differentiability}
K.~Tracy and Z.~Manchester, ``On the differentiability of the primal-dual interior-point method,'' 2024.

\bibitem[Montaut et~al.(2023)Montaut, Le~Lidec, et~al.]{montaut2023differentiable}
L.~Montaut, Q.~Le~Lidec \emph{et~al.}, ``Differentiable collision detection: a randomized smoothing approach,'' in \emph{2023 IEEE International Conference on Robotics and Automation (ICRA)}.\hskip 1em plus 0.5em minus 0.4em\relax IEEE, 2023, pp. 3240--3246.

\bibitem[Williams(1992)]{williams1992simple}
R.~J. Williams, ``Simple statistical gradient-following algorithms for connectionist reinforcement learning,'' \emph{Machine learning}, vol.~8, pp. 229--256, 1992.

\bibitem[Paulus et~al.(2025)Paulus, Geist, Schumacher, Musil, and Martius]{paulus2025hard}
A.~Paulus, A.~R. Geist, P.~Schumacher, V.~Musil, and G.~Martius, ``Hard contacts with soft gradients: Refining differentiable simulators for learning and control,'' \emph{arXiv preprint arXiv:2506.14186}, 2025.

\bibitem[Li et~al.(2020)Li, Ferguson, Schneider, Langlois, Zorin, Panozzo, Jiang, and Kaufman]{Li2020IPC}
M.~Li, Z.~Ferguson, T.~Schneider, T.~Langlois, D.~Zorin, D.~Panozzo, C.~Jiang, and D.~M. Kaufman, ``Incremental potential contact: Intersection- and inversion-free large deformation dynamics,'' \emph{ACM Trans. Graph. (SIGGRAPH)}, vol.~39, no.~4, 2020.

\bibitem[Paulus et~al.(2026)Paulus, Geist, Musil, Hoffmann, Beker, and Martius]{paulus2026softjax}
A.~Paulus, A.~R. Geist, V.~Musil, S.~Hoffmann, O.~Beker, and G.~Martius, ``{SoftJAX} \& {SoftTorch}: Empowering automatic differentiation libraries with informative gradients,'' \emph{arXiv preprint}, 2026.

\bibitem[Beker et~al.(2025)Beker, Gürtler, Shi, Geist, Razmjoo, Martius, and Calinon]{beker2025}
O.~Beker, N.~Gürtler, J.~Shi, R.~Geist, A.~Razmjoo, G.~Martius, and S.~Calinon, ``A smooth analytical formulation of collision detection and rigid body dynamics with contact,'' in \emph{2025 IEEE International Conference on Intelligent Robots and Systems (IROS)}.\hskip 1em plus 0.5em minus 0.4em\relax IEEE, 2025.

\bibitem[Beker et~al.(2026)Beker, Geist, Paulus, G{\"u}rtler, Shi, Calinon, and Martius]{beker2026smoothly}
O.~Beker, A.~R. Geist, A.~Paulus, N.~G{\"u}rtler, J.~Shi, S.~Calinon, and G.~Martius, ``Smoothly differentiable and efficiently vectorizable contact manifold generation,'' \emph{arXiv preprint arXiv:2602.20304}, 2026.

\bibitem[Liu et~al.(2023)Liu, Wu, Ruan, and Chirikjian]{liu2023marching}
W.~Liu, Y.~Wu, S.~Ruan, and G.~S. Chirikjian, ``Marching-primitives: Shape abstraction from signed distance function,'' in \emph{Proceedings of the IEEE/CVF Conference on Computer Vision and Pattern Recognition}, 2023, pp. 8771--8780.

\bibitem[Quilez(2013{\natexlab{a}})]{sdf_algebra}
\BIBentryALTinterwordspacing
I.~Quilez, \href{https://iquilezles.org/articles/distfunctions/}{``Signed distance functions},'' 2013.
\BIBentrySTDinterwordspacing

\bibitem[Quilez(2013{\natexlab{b}})]{smoothmin}
\BIBentryALTinterwordspacing
------, \href{https://iquilezles.org/articles/smin/}{``Smooth minimum},'' 2013.
\BIBentrySTDinterwordspacing

\bibitem[Spivak(1999)]{spivak_frenet_serret}
M.~Spivak, \emph{A Comprehensive Introduction to Differential Geometry}, 3rd~ed.\hskip 1em plus 0.5em minus 0.4em\relax Publish or Perish, Inc., 1999, vol.~2, ch. Curves In the Plane and In Space (The Serret--Frenet formulas), pp. 34--36.

\bibitem[Li and Calinon(2025)]{Li25RAL}
Y.~Li and S.~Calinon, ``From movement primitives to distance fields to dynamical systems,'' \emph{{IEEE} Robotics and Automation Letters ({RA-L})}, vol.~10, no.~9, pp. 9550--9556, 2025.

\bibitem[Artin(2013)]{artin2011algebra}
M.~Artin, \emph{Algebra}, 2nd~ed.\hskip 1em plus 0.5em minus 0.4em\relax Pearson, 2013, ch. 16 -- Galois Theory, pp. 501--502.

\bibitem[Habermann and Kindermann(2007)]{habermann2007multidimensional}
C.~Habermann and F.~Kindermann, ``Multidimensional spline interpolation: Theory and applications,'' \emph{Computational Economics}, vol.~30, no.~2, pp. 153--169, 2007.

\bibitem[Floater(2025)]{floater2023Spline}
\BIBentryALTinterwordspacing
M.~S. Floater, \href{https://www.uio.no/studier/emner/matnat/math/MAT4170/v25/undervisningsmateriale/spline_notes.pdf}{``An introduction to spline theory},'' Lecture Notes, Department of Mathematics, University of Oslo, 2025.
\BIBentrySTDinterwordspacing

\bibitem[Lai and Schumaker(2007)]{lai2007spline}
M.-J. Lai and L.~L. Schumaker, \emph{Spline functions on triangulations}.\hskip 1em plus 0.5em minus 0.4em\relax Cambridge University Press, 2007, no. 110.

\bibitem[Hart(1996)]{hart1996sphere}
J.~C. Hart, ``Sphere tracing: A geometric method for the antialiased ray tracing of implicit surfaces,'' \emph{The Visual Computer}, vol.~12, no.~10, pp. 527--545, 1996.

\bibitem[Crane(2005)]{crane2005ray}
K.~Crane, ``Ray tracing quaternion julia sets on the gpu,'' 2005.

\bibitem[Mordatch et~al.(2012)Mordatch, Todorov, and Popovi{\'c}]{mordatch2012discovery}
I.~Mordatch, E.~Todorov, and Z.~Popovi{\'c}, ``Discovery of complex behaviors through contact-invariant optimization,'' \emph{ACM Transactions on Graphics (ToG)}, vol.~31, no.~4, pp. 1--8, 2012.

\bibitem[Pang et~al.(2023)Pang, Suh, Yang, and Tedrake]{pang2023global}
T.~Pang, H.~T. Suh, L.~Yang, and R.~Tedrake, ``Global planning for contact-rich manipulation via local smoothing of quasi-dynamic contact models,'' \emph{IEEE Transactions on robotics}, 2023.

\bibitem[{NVIDIA Corporation}(2026)]{nvidia_cuda_guide}
{NVIDIA Corporation}, ``Cuda programming guide, release 13.2,'' \url{https://docs.nvidia.com/cuda/cuda-programming-guide/pdf/cuda-programming-guide.pdf}, 2026.

\bibitem[Morton(1966)]{morton1966computer}
G.~M. Morton, ``A computer oriented geodetic data base and a new technique in file sequencing.(1966),'' 1966.

\bibitem[Wang et~al.(2017)Wang, Liu, Guo, Sun, and Tong]{wang2017ocnn}
P.-S. Wang, Y.~Liu, Y.-X. Guo, C.-Y. Sun, and X.~Tong, ``{O-CNN}: Octree-based convolutional neural networksfor {3D} shape analysis,'' \emph{ACM Transactions on Graphics (SIGGRAPH)}, vol.~36, no.~4, 2017.

\bibitem[Wu et~al.(2024)Wu, Jiang, Wang, Liu, Liu, Qiao, Ouyang, He, and Zhao]{wu2024ptv3}
X.~Wu, L.~Jiang, P.-S. Wang, Z.~Liu, X.~Liu, Y.~Qiao, W.~Ouyang, T.~He, and H.~Zhao, ``Point transformer v3: Simpler, faster, stronger,'' in \emph{CVPR}, 2024.

\bibitem[Kurtz and Castro(2025)]{kurtz2025cenic}
V.~Kurtz and A.~Castro, ``Cenic: Convex error-controlled numerical integration for contact,'' \emph{arXiv preprint arXiv:2511.08771}, 2025.

\bibitem[Kurtz et~al.(2026)Kurtz, Castro, {\"O}nol, and Lin]{kurtz2026inverse}
V.~Kurtz, A.~Castro, A.~{\"O}. {\"O}nol, and H.~Lin, ``Inverse dynamics trajectory optimization for contact-implicit model predictive control,'' \emph{The International Journal of Robotics Research}, vol.~45, no.~1, pp. 23--40, 2026.

\bibitem[Elandt et~al.(2019)Elandt, Drumwright, Sherman, and Ruina]{hydroelastic}
\BIBentryALTinterwordspacing
R.~Elandt, E.~Drumwright, M.~Sherman, and A.~Ruina, \href{https://doi.org/10.1109/IROS40897.2019.8968548}{``A pressure field model for fast, robust approximation of net contact force and moment between nominally rigid objects},'' in \emph{2019 IEEE/RSJ International Conference on Intelligent Robots and Systems (IROS)}.\hskip 1em plus 0.5em minus 0.4em\relax IEEE Press, 2019, p. 8238–8245.
\BIBentrySTDinterwordspacing

\bibitem[Johnson(1987)]{johnson1987contact}
K.~L. Johnson, \emph{Contact mechanics}.\hskip 1em plus 0.5em minus 0.4em\relax Cambridge university press, 1987.

\bibitem[Hunt and Crossley(1975)]{hunt1975coefficient}
K.~H. Hunt and F.~R. Crossley, ``Coefficient of restitution interpreted as damping in vibroimpact,'' 1975.

\bibitem[Corporation()]{mscadams}
\BIBentryALTinterwordspacing
M.~S. Corporation, \href{https://hexagon.com/products/product-groups/computer-aided-engineering-software/adams}{``Adams (automated dynamic analysis of mechanical systems)}.''
\BIBentrySTDinterwordspacing

\bibitem[Macklin(2022)]{Macklin_Warp_A_High-performance_2022}
\BIBentryALTinterwordspacing
M.~Macklin, \href{https://github.com/NVIDIA/warp}{``{Warp: A High-performance Python Framework for GPU Simulation and Graphics}},'' Mar. 2022, nVIDIA GPU Technology Conference (GTC).
\BIBentrySTDinterwordspacing

\end{thebibliography}
